\documentclass[11pt]{article}

\usepackage[final]{acl}

\usepackage{times}
\usepackage{latexsym}

\usepackage[T1]{fontenc}

\usepackage[utf8]{inputenc}

\usepackage{microtype}

\usepackage{inconsolata}
\usepackage{enumitem}
\usepackage{graphicx}
\usepackage{derivative}
\usepackage{amsmath}
\usepackage[nameinlink]{cleveref}
\usepackage{mathtools}
\usepackage{commath}
\usepackage{scalerel}
\usepackage{multirow}
\usepackage{pgfplots}
\usepackage{algorithm}
\usepackage{algorithmic}
\usepackage{url}            
\usepackage{booktabs}       
\usepackage{amsfonts}       
\usepackage{nicefrac}       
\usepackage{microtype}      
\usepackage{xcolor}         
\usepackage{subcaption}

\usepackage{graphicx}
\AtBeginDocument{%
  }
\newcommand\model{Syn2DRE}

\title{Syn2DRE: Synergistic Dual Teacher-Student Framework for Document-level Relation Extraction}

\author{
     \textbf{Khai P. Tran},
     \textbf{Xue Li}
    \\
    \\
     School of Electrical Engineering and Computer Science \\ The University of Queensland, Australia \\
    \\
     \small{
       \textbf{Correspondence:} \href{mailto:phankhai.tran@uq.edu.au}{phankhai.tran@uq.edu.au}
     }
}

\begin{document}
\maketitle
\begin{abstract}
    Document-level Relation Extraction (DocRE) aims to identify relations between entities across a document but typically relies on costly large-scale annotations. While pseudo-labelling enables the use of low-quality data, existing methods often suffer from inefficiency and confirmation bias. To address these challenges, we propose \model, a semi-supervised framework that combines self-supervised learning with cross-supervision in a lightweight teacher–student architecture to produce more reliable pseudo-labels. Additionally, we introduce a unified end-to-end training strategy to further improve efficiency of the framework. Experiments on multiple benchmarks show that \model\ consistently outperforms prior state-of-the-art semi-supervised DocRE methods. Our code is released at: {\tt \href{https://github.com/khaitran22/Syn2DRE}{https://github.com/khaitran22/Syn2DRE}}

\end{abstract}

\section{Introduction}
Relation Extraction (RE) identifies semantic relationships between entity pairs and is fundamental to applications such as Question Answering \cite{xu2016question}. Early work focuses on sentence-level contexts, but many real-world relations span multiple sentences; for example, over 40\% of relational facts in Wikipedia require cross-sentence reasoning \cite{yao2019docred}. This limitation has led to growing interest in Document-level Relation Extraction (DocRE) \cite{xu2021document, zhang2021document, zhou2021document, tan2022document}, which models relations across entire documents.

DocRE methods typically rely on deep neural networks, which require substantial amounts of annotated data for effective training. However, creating high-quality DocRE datasets is both time-consuming and error-prone \cite{tan2022revisiting, huang2022does}. For example, the widely-used DocRED dataset \cite{yao2019docred} contains an average of 19.5 entities per document, resulting in roughly 38,000 candidate triples to be annotated per document when considering 97 predefined relation types \cite{tan2022revisiting}.

To reduce reliance on manual annotation, several studies \cite{tan2023class, feng2024dure, sun2023uncertainty} have explored pseudo-labelling \cite{lee2013pseudo} -- an approach that generates synthetic labels for unlabelled or noisily labelled data. This typically involves an iterative loop: generating pseudo-labels, incorporating them into the training set, retraining the model, and repeating the cycle. However, such methods are prone to confirmation bias \cite{arazo2020pseudo}, where low-quality pseudo-labels are reinforced over iterations, harming overall performance. Some recent works \cite{tan2022document, tan2023class, feng2024dure} use pseudo-labels without evaluating their reliability, further exacerbating this issue. To address this, \citet{sun2023uncertainty} proposed uncertainty-based selection to improve pseudo-label quality. While effective, their method relies on a single model and requires multiple rounds of training and denoising, significantly increasing computational cost and leaving confirmation bias largely unaddressed.

In this work, we introduce the \textbf{Synergistic Dual Teacher-Student} approach (\model), a novel semi-supervised learning framework for DocRE that generates high-quality pseudo-labels in a single training run, eliminating the need for iterative denoising. To reduce training overhead, the teacher model is derived from the student using Exponential Moving Average (EMA) to produce more stable targets \cite{tarvainen2017mean}.

While EMA-based pseudo-label generation enhances performance, it may still suffer from confirmation bias. To address this issue, we propose an unsupervised synergistic cross-supervision strategy inspired by Co-teaching \cite{blum1998combining, han2018co, yu2019does}. Specifically, two distinct teacher-student pairs exchange highly confident pseudo-labels on instances where their predictions diverge, promoting diverse learning signals and mitigating confirmation bias. Confidence is determined by evaluating the variance of prediction logits across multiple forward passes. Finally, we present a unified training framework that integrates both self-supervised learning and cross-supervision cohesively.

We evaluate \model\ on three general-domain DocRE benchmark datasets -- DocRED \cite{yao2019docred}, Re-DocRED \cite{tan2022revisiting}, and HacRED \cite{cheng2021hacred} -- as well as two biomedical datasets, CDR \cite{li2016biocreative} and GDA \cite{wu2019renet}. Our framework consistently achieves state-of-the-art performance across these datasets.

The main contributions of this work are as follows:
\begin{itemize}[noitemsep]
    \item \textbf{Novel semi-supervised DocRE framework:} We propose \model, a dual teacher-student framework that leverages both unlabelled and noisy data to boost DocRE performance.
    \item \textbf{Self-supervised with Unsupervised Synergistic Cross-supervision:} \model\ combines EMA-based Pseudo-labelling step with Unsupervised Synergistic Cross-supervision strategy to reduce confirmation bias -- marking the first such integration in DocRE.
    \item \textbf{Unified training process:} A unified training procedure integrates both steps in an end-to-end manner to enhance performance in DocRE.
    \item \textbf{Superior empirical results:} Extensive experiments on popular DocRE datasets show that \model\ significantly outperforms existing semi-supervised baselines on standard DocRE benchmarks.
\end{itemize}

\begin{figure*}[!t]
    \centering
    \includegraphics[width=\textwidth]{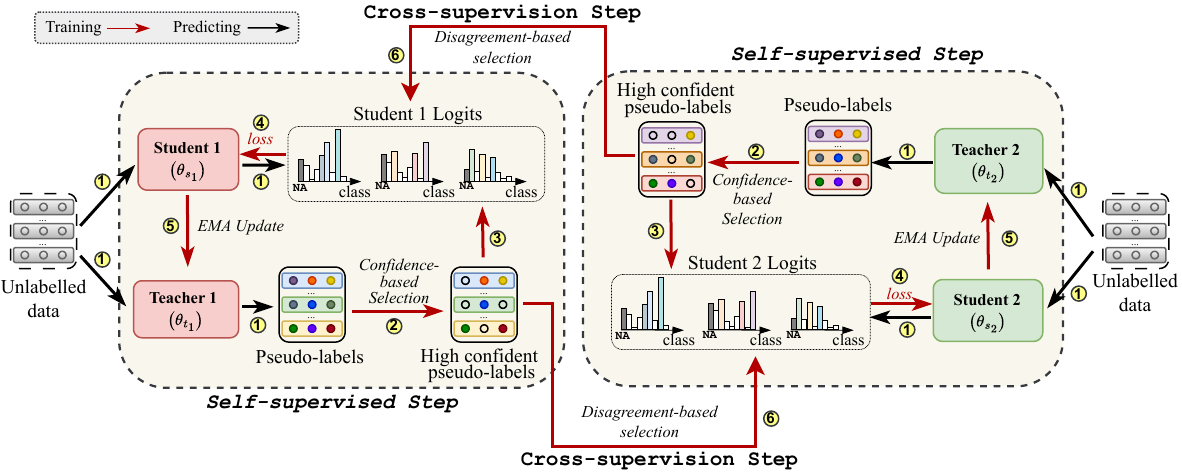}
    \caption{An overview of the proposed \model\ method.}
    \label{fig:overall-framework}
\end{figure*}
\section{Problem Formulation}
Let $D = \{w_l\}_{l=1}^{L}$ be a document with $L$ words and a set of entities $\mathcal{E}_D = \{e_i\}_{i=1}^{|\mathcal{E}_D|}$. Each entity $e_i$ has a set of mentions $\mathcal{M}_{e_i} = \{m^i_j\}_{j=1}^{|\mathcal{M}_{e_i}|}$, i.e., phrases referring to the same entity. In the DocRE task, the goal is to predict a subset of relations from a predefined set $\mathcal{R} = \{r_k\}_{k=1}^{|\mathcal{R}|}$ that hold between entity pairs $(e_h, e_t)$, where $h \neq t$. For brevity, we denote such pairs as $(h, t)$. A relation $r \in \mathcal{R}$ is considered present between $e_h$ and $e_t$ if it is expressed between any of their mentions; these are labelled as positive pairs. If no relation exists between any mention pair, the entity pair is labelled {\tt NA} (i.e., negative) relation. Note that positive pairs may express multiple relations in $\mathcal{R}$.
\section{Methodology}
\subsection{Framework Overview}
\paragraph{Motivation.}
We propose a robust semi-supervised learning framework for the DocRE task, which leverages the abundant unlabelled data for model training. Our approach is grounded in the following three key observations in the DocRE task: (1) obtaining large-scale, high-quality labelled data is prohibitively expensive; (2) the potential of unlabelled data to enhance performance remains under-explored; and (3) existing pseudo-labelling methods often fail to assess label quality or rely on iterative processes that can amplify noise, ultimately degrading model performance.

\paragraph{}
To address these challenges, we introduce \model, a novel framework featuring dual teacher-student networks trained in parallel. As shown in Figure~\ref{fig:overall-framework}, during the Self-supervised step, teacher networks generate pseudo-labels for unlabelled instances, while student networks make predictions on the same data (step \raisebox{-0.2\baselineskip}{\includegraphics[height=1em]{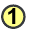}}). Pseudo-labels with high confidence are selected (step \raisebox{-0.2\baselineskip}{\includegraphics[height=1em]{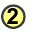}}) and used to update the corresponding student networks (steps \raisebox{-0.2\baselineskip}{\includegraphics[height=1em]{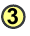}}, \raisebox{-0.2\baselineskip}{\includegraphics[height=1em]{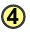}}). The improved associated student networks then inform updates to the teacher networks (step \raisebox{-0.2\baselineskip}{\includegraphics[height=1em]{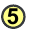}}). To reduce confirmation bias, we periodically apply the proposed unsupervised synergistic cross-supervision mechanism to further refine the opposite student models (step \raisebox{-0.2\baselineskip}{\includegraphics[height=1em]{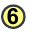}}).

\subsection{Teacher-Student Architectures}
\label{sec:baseline-models}

\subsubsection{Student Networks.}
\label{subsec:student-nets}
To promote complementary learning signals during cross-supervision, we adopt KD-DocRE \citep{tan2022document} and UGDRE \citep{sun2023uncertainty} as our student networks. These lightweight models differ in both entity pair representation and training objectives. KD-DocRE employs Axial Attention \citep{ho2019axial} and a composite loss function that integrates Adaptive Thresholding Loss (ATL) \citep{zhou2021document} with Focal Loss \citep{lin2017focal} to better handle long-tail relations. In contrast, UGDRE enhances entity representations through token-level graph construction and follows the original ATL objective. Full architectural details are provided in Appendix \ref{appendix:baseline-structures}.

\subsubsection{Teacher Networks.}
Each teacher network in \model\ mirrors the architecture of its associated student. Their parameters are updated using EMA, which smooths the weights over time to yield more stable pseudo-labels. Initialisation details are discussed in Section \ref{subsec:train-procedure}.

\subsection{Self-supervised Step}
\label{subsec:self-training}
This section details the Self-supervised process. Pseudo-label generation and refinement (steps \raisebox{-0.2\baselineskip}{\includegraphics[height=1em]{figs/1.pdf}} and \raisebox{-0.2\baselineskip}{\includegraphics[height=1em]{figs/2.pdf}}) are described in Sections~\ref{sec:pseudo-label-gen} and \ref{sec:pseudo-label-refine}, while the teacher update mechanism (step \raisebox{-0.2\baselineskip}{\includegraphics[height=1em]{figs/5.pdf}}) is presented in Section~\ref{sec:teacher-net-update}.

\paragraph{Motivation.} The Self-supervised step in \model\ aims to reduce the prolonged training time of prior methods by generating high-quality pseudo-labels in a single pass via teacher-student networks, avoiding iterative denoising. 

\subsubsection{Pseudo-label Generation}
\label{sec:pseudo-label-gen}
In the Self-supervised step, each teacher network generates pseudo-labels for unlabelled data to supervise its associated student. To ensure label reliability, we adopt Monte Carlo (MC) dropout \citep{gal2016dropout}. Specifically, we perform several stochastic forward passes to compute prediction logits for each relation $r_i \in \mathcal{R}$ and the threshold class $TH$ \citep{zhou2021document}, representing the {\tt NA} relation, for a given entity pair $(e_h, e_t)$. The final prediction score $\bar{sc}_{h,t}^{r_i}$ is obtained by averaging over those forward passes:
\begin{equation}
    \label{pred-logits}
    \bar{sc}_{h,t}^{r_i} = \frac{1}{N} \sum_{j=1}^N sc_{h,t}^{r_{ij}}
\end{equation}
where $sc_{h,t}^{r_{ij}}$ is the logit score for relation $r_i$ in the $j^{th}$ pass, $N$ is the number of stochastic forward passes. The same averaging is applied to the $TH$ class.

Pseudo-labels are then assigned according to the following rule:
$$
\text{Pseudo-label} =
\begin{cases}
    r_i & \text{if } \bar{sc}_{h,t}^{r_i} > \bar{sc}_{h,t}^{TH} \\
    \texttt{NA} & \text{otherwise}
\end{cases}
$$
where $\bar{sc}_{h,t}^{TH}$ denotes the averaged score for the $TH$ class.


\subsubsection{Pseudo-labels Refinement}
\label{sec:pseudo-label-refine}

Neural networks often exhibit poor calibration, producing overconfident yet incorrect pseudo-labels \citep{guo2017calibration, rizve2021in, si2024improving}. To address this, we propose an unsupervised refinement strategy that prioritises reliable pseudo-labels using uncertainty,

\paragraph{Uncertainty Estimation.} The uncertainty $u^{r_i}_{h,t}$ for each pseudo-relation $r_i$ of an entity pair $(e_h, e_t)$ is quantified as the variance of its logits across multiple forward passes, defined as:

\begin{equation}
\label{uncertainty-score}
u^{r_i}_{h,t} = \sqrt{\frac{\sum_{j=1}^N(sc_{h,t}^{r_{ij}} - \bar{sc}_{h,t}^{r_i})^2}{N}}
\end{equation}

Here, $\bar{sc}_{h,t}^{r_i}$ is the mean prediction logit across $N$ stochastic forward passes. We compute uncertainty for all entity pairs, including both positive and negative ones. If a positive pair $(e_h', e_t')$ has multiple pseudo-relations, its overall uncertainty is averaged across these relations:
\begin{equation}
u_{h',t'} = \frac{1}{|\mathcal{R}^{ps}_{h', t'}|} \sum_{i=1}^{|\mathcal{R}^{ps}_{h', t'}|} u^{r_i}_{h',t'}
\end{equation}
where $\mathcal{R}^{ps}_{h', t'} \subset \mathcal{R}$ denotes the set of pseudo-relations. We rank all positive pairs by increasing uncertainty and select the top $p\%$ as reliable training samples; likewise, we select the top $n\%$ negative pairs. To address the class imbalance issue -- where 97\% of entity pairs are negative -- we set $p = 30$ and $n = 10$.

The selected pseudo-labels are then used to update the student networks with standard training process.

\subsubsection{Teacher Network Update}
\label{sec:teacher-net-update}
Once the student networks are trained, we update the teacher models using EMA of student parameters, as follows:
\begin{equation}
    \label{eq:ema-teacher-update}
    \theta_{t}^{(j)} \leftarrow \alpha \cdot \theta_t^{(j-1)} + (1-\alpha) \cdot \theta_s^{(j)}
\end{equation}

Here, $\theta_{t}^{(j)}$ and $\theta_s^{(j)}$ denote the teacher and student parameters at iteration $j$, and $\alpha$ is the smoothing factor. This EMA update can also be reformulated as:

\begin{equation}
\label{ema-backprop}
\theta_{t}^{(j)} = \theta_{t}^{(j-1)} - \gamma(1-\alpha)\sum_{t=0}^{j-1}\alpha^{j-1-t}\pdv{\mathcal{L}}{{\theta_s^{(t)}}}
\end{equation}
\noindent
where $\gamma$ is the learning rate. As $\alpha \approx 1$ (e.g., 0.995), updates incorporate small, weighted student gradients, forming a conservative, ensemble-like strategy that reduces bias and improves pseudo-label reliability \citep{zhang2021improving}. Derivation of the Eq. (\ref{ema-backprop}) is detailed in the Appendix \ref{sec:eq-5-derivation}.

\subsection{Unsupervised Synergistic Cross-supervision Step}
\paragraph{Motivation.}
In the Self-supervised step, teacher-generated pseudo-labels help students improve. However, using the same architecture for both models may perpetuate confirmation bias, leading to sub-optimal performance. To address this, we introduce an Unsupervised Synergistic Cross-supervision training strategy (step \raisebox{-0.2\baselineskip}{\includegraphics[height=1em]{figs/6.pdf}}).

To reduce confirmation bias, a model should learn from signals it cannot easily infer on its own. Motivated by this, we co-train two DocRE models with different architectures that teach each other, leveraging complementary graph-based and transformer-based representations and distinct objectives for class imbalance, as described in $\S$ \ref{sec:baseline-models}.

\subsubsection{Procedure}
During training, each teacher reuses samples with the lowest uncertainty, as defined in Section \ref{sec:pseudo-label-refine}. From these, disagreeing pseudo-labels are exchanged, such that teacher 2 supervises student 1 and teacher 1 supervises student 2.

Formally, let $(t_1, s_1)$ and $(t_2, s_2)$ denote the teacher-student pairs. We extract subsets $\hat{D}_{t_1}$ and $\hat{D}_{t_2}$ containing the pseudo-labelled samples from each teacher with the lowest uncertainty. From these, we identify the sets of disagreed samples $\tilde{D}_{t_1}$ and $\tilde{D}_{t_2}$, where:
\begin{equation*}
    \begin{split}
        \tilde{D}_{t_2} &= \hat{D}_{t_2} \setminus \hat{D}_{t_1}, \\
        \tilde{D}_{t_1} &= \hat{D}_{t_1} \setminus \hat{D}_{t_2}
    \end{split}
\end{equation*}

\subsubsection{Objective.} The associated soft pseudo-labels $\tilde{l}_{t_1}, \tilde{l}_{t_2}$ of $\tilde{D}_{t_1}, \tilde{D}_{t_2}$ are used to supervise the opposite students. We adopt soft labels (probability distributions over $\mathcal{R} \cup \texttt{NA}$) to preserve uncertainty and ensure stable training. The cross-supervision loss $\mathcal{L}_{css}$ is defined as:

\begin{equation}
\label{eq:cross-stage-loss}
\mathcal{L}_{css} = \text{MeanSquaredError}\left(\tilde{l}_{s_i}, \tilde{l}_{t_j}\right)
\end{equation}
\noindent
where $i, j \in \{1,2\}$ and $i \neq j$ and $\tilde{l}_s, \tilde{l}_t$ denote the predictive distributions from the student and teacher, respectively.

\subsection{Unified Training Procedure}
\label{subsec:train-procedure}
We propose a unified training procedure to improve the framework’s efficiency, consisting of two phases: (1) pre-training and (2) fine-tuning.

\subsubsection{Pre-training Phase}
\label{sec:pre-training-stage}
During the pre-training phase, we initialise the student networks $s_1$ and $s_2$ using distantly supervised DocRE dataset with standard training process to improve generalisation. After pre-training, we copy their parameters to the respective teacher networks to facilitate a strong initialisation for both teacher and student networks, as follows:
\begin{equation}
    \label{assign-params}
    \begin{aligned}
        \theta_{t_1} &\leftarrow \theta_{s_1} \\
        \theta_{t_2} &\leftarrow \theta_{s_2}
    \end{aligned}
\end{equation}

\subsubsection{Fine-tuning Phase}
\label{sec:fine-tuning-stage}

In the fine-tuning phase, the teacher-student pairs alternate between the Self-supervised step and the Unsupervised Synergistic Cross-supervision step. To ensure robust initialisation for these steps, we first perform a warm-up round using human-annotated data for $n$ epochs.

After warming up, the Self-supervised step begins with refined pseudo-labels, and the Unsupervised Synergistic Cross-supervision step is conducted periodically every $m$ steps to avoid mutual overfitting due to residual label noise. The overall loss $\mathcal{L}_{ft}$ for the this step for each student network is:

\begin{equation}
\label{eq:finetune-loss}
\mathcal{L}_{ft} = \mathcal{L}_{st} + \lambda_{ul} \cdot \mathcal{L}_{css}
\end{equation}
\noindent
where $\mathcal{L}_{st}$ is the loss value of the Self-supervised Step ($\S$\ref{subsec:self-training}) and the hyper-parameter $\lambda_{ul}$ controls the contribution of $\mathcal{L}_{css}$.

To further combat confirmation bias, we also include supervised loss on human-annotated data $\mathcal{L}_{sp}$ for each student network. The final training objective $\mathcal{L}_{fn}$ becomes:
\begin{equation}
    \label{eq:total-loss}
    \mathcal{L}_{fn} = \mathcal{L}_{sp} + \lambda_2 \cdot \mathcal{L}_{ft}
\end{equation}
where $\lambda_2$ controls the contribution of $\mathcal{L}_{ft}$.
\begin{table*}[!t]
    \centering
    \resizebox{\textwidth}{!}{
    \begin{tabular}{l c c c c c c c c}
        \toprule
        \multirow{3}*{\textbf{Model}} & \multicolumn{4}{c}{\textbf{DocRED}} & \multicolumn{4}{c}{\textbf{Re-DocRED}} \\
        \cmidrule(r){2-5}
        \cmidrule(r){6-9}
         &\multicolumn{2}{c}{\textbf{Dev}} & \multicolumn{2}{c}{\textbf{Test}}&\multicolumn{2}{c}{\textbf{Dev}} & \multicolumn{2}{c}{\textbf{Test}}\\
        \cmidrule(r){2-3}
        \cmidrule(r){4-5}
        \cmidrule(r){6-7}
        \cmidrule(r){8-9}
         & $F_1$ & Ign $F_1$ & $F_1$ & Ign $F_1$ & $F_1$ & Ign $F_1$ & $F_1$ & Ign $F_1$ \\
        \midrule
        
        ATLOP \citep{zhou2021document}          & 63.42 & 61.57 & 63.48 & 61.43 & 74.34 & 73.62 & 74.23 & 73.53 \\
        DocuNet \citep{zhang2021document}       & 64.35 & 62.66 & 64.00 & 61.93 & 76.22 & 75.50 & 75.35 & 74.61 \\
        NCRL \citep{zhou2022none}               & 63.87 & 61.65 & 63.45 & 60.98 & 75.85 & 74.91 & 75.90 & 75.00 \\
        CAST* \citep{tan2023class}              & -     & -     & -     & -     & -     & -     & 74.67 & 73.32 \\
        VaeDiff-DocRE* \citep{tran2025vaediff}   & -     & -     & -     & -     & 75.89 & 74.96 & 75.07 & 74.13 \\
        SSR-PU \citep{wang2022unified}          & 63.00 & 60.43 & 62.66 & 59.80 & 76.83 & 75.57 & 76.23 & 74.96 \\
        KD-NA \citep{tan2022document}           & 64.17 & 62.18 & 64.12 & 61.77 & 76.14 & 75.25 & 76.00 & 75.12 \\
        KD-DocRE \citep{tan2022document}        & 64.81 & 62.62 & 64.76 & 62.56 & 76.47 & 75.30 & 76.14 & 74.97 \\
        DURE* \citep{feng2024dure}              & -     & -     & -     & -     & -     & -     & 76.84 & 75.32 \\
        UGDRE \citep{sun2023uncertainty}         & 65.71 & 63.62 & 65.58 & 63.26 & 78.28 & 77.32 & 78.14 & 77.24 \\
        \midrule
        \textbf{\model \, (Ours)}               & \textbf{66.09\small{$\pm$0.17}} & \textbf{63.88\small{$\pm$0.2}} & \textbf{66.14}\textsuperscript{\textdagger} & \textbf{63.66}\textsuperscript{\textdagger} & \textbf{79.53\small{$\pm$0.21}} & \textbf{78.28\small{$\pm$0.23}} & \textbf{79.40\small{$\pm$0.22}} & \textbf{78.16\small{$\pm$0.22}}\\
        \bottomrule
    \end{tabular}}
    \caption{Experimental results on DocRED and Re-DocRED datasets. Baseline results are from \citet{sun2023uncertainty}. \textsuperscript{\textdagger} indicates the best single run; * denotes results reported directly from original papers.}
    \label{tab:exp-results}
\end{table*}

\section{Experiments}
In this section, we present a series of experiments to assess the effectiveness of our proposed framework. We begin by outlining the experimental setup, followed by empirical analyses aimed at addressing the following research questions (RQs):
\begin{itemize}[noitemsep]
    \item \textbf{RQ1:} How effective is the overall framework in different datasets?
    \item \textbf{RQ2:} How well does the method handle long-tailed relation distributions?
    \item \textbf{RQ3:} How effective is each of the proposed components?
    \item \textbf{RQ4:} How resource-efficient is the proposed approach compared to existing baselines?
    \item \textbf{RQ5:} How is \model\ effective when using with other DocRE architectures?
\end{itemize}

\subsection{Datasets}
\label{subsec:experimental-setup}
We conducted evaluations on five widely used publicly available benchmarks for the DocRE task. More details are shown in Table \ref{tab:datasets-statistics} in Appendix \ref{appendix:experimental-details}.

\noindent
\textbf{DocRED.} \ DocRED contains 96 predefined relation types, curated from Wikipedia and Wikidata. It includes 3,053 training, 1,000 validation, and 1,000 test documents. A distantly supervised version with 101,873 documents was also released by \citet{yao2019docred}.

\noindent
\textbf{Re-DocRED.} \ Re-DocRED is a refined variant of DocRED that addresses annotation incompleteness by adding previously overlooked relational facts. It consists of 3,053 training, 500 development, and 500 test documents.

\noindent
\textbf{HacRED.} \ HacRED is a Chinese document-level relation extraction dataset designed to address challenging cases. It defines 27 difficult relation types and is split into 6,231/1,500/1,500 instances for training, validation, and test sets.

\noindent
\textbf{CDR.} \ CDR is a human-annotated biomedical dataset containing 500 training documents. The task is to predict binary interactions between chemical and disease entities.

\noindent
\textbf{GDA.} \ GDA is a large-scale biomedical dataset comprising 29,192 training articles. The task is to predict binary interactions between gene and disease entities. Dataset splits follow \citet{christopoulou2019connecting} and \citet{zhou2021document}. \\

\noindent
Since HacRED, CDR, and GDA do not provide unlabelled data, we treat their validation sets as unlabelled and avoid pretraining on their original labels to ensure a fair evaluation.

\subsection{Experiment Setup}
\subsubsection{Evaluation Metrics}
We adopt the micro-averaged $\mathbf{F_1}$ score as the primary evaluation metric across all five datasets. Following \citet{yao2019docred}, we also report the \textbf{Ignored} $\mathbf{F_1}$, which omits relational facts overlapping between the training and test sets for the DocRED and Re-DocRED datasets. Additionally, to assess model performance across relation frequency levels in Re-DocRED, we include \textbf{Freq} $\mathbf{F_1}$ (for the 10 most frequent relations) and \textbf{LT} $\mathbf{F_1}$ (for the remaining 86), as introduced by \citet{tan2022revisiting}.

\subsubsection{Baselines}
We compare \model\, with several competitive baselines that generate pseudo-labels to augment human-labeled data. \textit{KD-DocRE} \citep{tan2022document} leverages knowledge distillation to produce soft pseudo-labels. \textit{UGDRE} \citep{sun2023uncertainty} uses uncertainty-aware, multi-phase training to produce hard pseudo-labels. \textit{VaeDiff-DocRE} \citep{tran2025vaediff} creates pseudo-representations for entity pairs using data augmentation. \textit{DURE} \citep{feng2024dure} generates pseudo-labels for unlabelled data via controllable text generation conditioned on relation triplets.

\subsubsection{Implementation Details}
We use BERT-base-cased \citep{devlin2019bert}, XLM-RoBERTa-Large \cite{conneau2020unsupervised}, and SciBERT \cite{beltagy2019scibert} as encoders, and train all models with AdamW \citep{loshchilov2019decoupled}, adopting official implementations for baselines. Pre-training follows original hyperparameters (Appendix \ref{appendix:baseline-structures}), while in fine-tuning, EMA rates are set to 0.995 (UGDRE) and 0.999 (KD-DocRE), starting after 10 steps and updating every 5 steps, with unchanged learning rates. Fine-tuning runs for 25 epochs on Re-DocRED and 7 on DocRED, sampling 16 unlabelled documents per step, with warm-up lasting 15 and 5 epochs, respectively. For cross-supervision, we set $m = 6$, $\lambda_{ul} = 0.1$, and $\lambda_2 = 0.5$. Our experiments were conducted on a server equipped with an NVIDIA H100 GPU (80GB).

\subsection{Overall Performance}
\subsubsection{General-domain Datasets}
\paragraph{DocRED and Re-DocRED.}
To address \textbf{RQ1}, we evaluate \model\ on DocRED and Re-DocRED using the student architectures described in $\S$\ref{subsec:student-nets}, with results in Table~\ref{tab:exp-results}; all improvements are statistically significant ($p < 0.01$) under the paired test of \citet{dror2018hitchhiker}. Results with other student architectures are reported in Section \ref{subsec:result-diff-models}.



\model\ significantly outperforms all baselines, particularly pseudo-labelling approaches, while also demonstrating greater robustness than augmentation-based methods. Compared to UGDRE, our method achieves consistent gains of +0.38/+0.56 F1 on the DocRED development/test sets, and larger improvements over KD-DocRE of +1.28/+1.38 F1. On Re-DocRED, \model\ surpasses UGDRE by +1.25/+1.26 F1 and KD-DocRE by +3.06/+3.26 F1 (dev/test). In addition, \model\ achieves notable gains over augmentation-based baselines, including +2.56 F1 over DURE (test) and +3.64/+4.33 F1 over VaeDiff-DocRE (dev/test), highlighting the effectiveness and robustness of our pseudo-label-driven strategy without relying on explicit text or embedding synthesis.

\paragraph{HacRED.}
From Table \ref{tab:hacred-perf}, \model\ significantly outperforms state-of-the-art baselines on challenging non-English cases. In particular, \model\ achieves average F1 gains of +0.61 and +0.23 over the baselines. These improvements could be further amplified with additional unlabelled documents, which are inexpensive to obtain in practice.
\begin{table}[!ht]
    \centering
    \resizebox{0.78\linewidth}{!}{%
    \begin{tabular}{lc}
    \toprule
    \textbf{Model}                                 & \multicolumn{1}{c}{\textbf{Test F1}} \\ \midrule
    SSAN \cite{xu2021entity}                       & 61.00                           \\
    LSR \cite{nan2020reasoning}                    & 68.41                           \\
    GAIN \cite{zeng2020double}                     & 76.09                           \\
    ATLOP \cite{zhou2021document}                  & 77.63                           \\
    KD-DocRE \cite{tan2022document}                & 78.75                           \\
    DAATF+ATL \cite{wei2025document}               & 79.03                           \\
    DAATF+AFL \cite{wei2025document}               & 79.42                           \\ \midrule
    \textbf{\model\ (Ours)}                         & \textbf{79.64\small{$\pm$0.1}}                  \\ \bottomrule
    \end{tabular}%
    }
    \caption{Experimental results on HacRED dataset. Baseline results are from \citet{wei2025document}.}
    \label{tab:hacred-perf}
    \vspace{-5mm}
\end{table}

\subsubsection{Biomedical Datasets}
To evaluate our framework in the biomedical domain, we conducted experiments on the CDR and GDA datasets using ATLOP and DocuNet as baselines, with results reported in Table \ref{tab:cdr-gda-perf}.
\begin{table}[!ht]
\centering
\resizebox{0.8\columnwidth}{!}{%
\begin{tabular}{lcc}
\toprule
\multirow{2}{*}{\textbf{Model}} & \multicolumn{2}{c}{\textbf{Test F1}} \\ \cmidrule(l){2-3} 
                                & \textit{CDR}      & \textit{GDA}     \\ \midrule
SSAN \cite{xu2021entity}                        & 68.7                  & 83.7                      \\
ATLOP \cite{zhou2021document}                   & 69.4                  & 83.9                      \\
DGI \cite{wang2023denoising}                    & 70.1                  & 84.1                      \\
SE-REPI \cite{liu2025document}                  & 73.3                  & 84.8                      \\
DAATF \cite{wei2025document}                    & 71.62                 & 85.04                     \\
DocuNet \cite{zhang2021document}                & 76.3                  & 85.3                      \\ \midrule
\textbf{\model\ (Ours)}                         & \textbf{84.77}        & \textbf{85.45}            \\ \bottomrule
\end{tabular}%
}
\caption{Model performance on CDR and GDA datasets.}
\label{tab:cdr-gda-perf}
\vspace{-5mm}
\end{table}

As shown in the table, \model\ achieves state-of-the-art performance on both biomedical DocRE datasets. Notably, it improves the F1 score on the CDR dataset by 8.47 points over the best baseline. On the GDA dataset, a smaller yet consistent improvement of +0.15 in F1 score is also observed.

\subsection{Performance on Long-tailed Relations}
To address \textbf{RQ2}, we evaluate \model\ on rare relations in Re-DocRED (Table~\ref{tab:long-tailed-result}). \model\ outperforms generation-based baselines, achieving LT $F_1$ gains of +2.02 over \textit{VaeDiff-DocRE} and +3.78 over \textit{DuRE}. These results show that \model\ robustly handles infrequent relations by generating high-quality pseudo-labels without explicit text generation, enabling scalable data augmentation for improved DocRE performance.

\begin{table}[!ht]
\centering
    \resizebox{\linewidth}{!}{%
    \begin{tabular}{lcc}
        \toprule
        \textbf{Model}                           & \textbf{Freq $\mathbf{F_1}$}           & \textbf{LT $\mathbf{F_1}$}            \\ \midrule
        ATLOP \citep{zhou2021document}           & 75.92                                  & 67.46                                 \\
        SSR-PU \citep{wang2022unified}           & 78.41                                  & 68.32                                 \\
        CREST \citep{wei2021theoretical}         & 77.93                                  & 68.68                                 \\
        CAST \citep{tan2023class}                & 78.53                                  & 69.34                                 \\
        VaeDiff-DocRE \citep{tran2025vaediff}    & 79.18                                  & 71.12                                 \\
        DuRE \citep{feng2024dure}                & 79.81                                  & 72.88                                 \\ \midrule
        \textbf{\model}                          & \textbf{82.42\small{$\pm$0.26}}        & \textbf{74.90\small{$\pm$0.23}}       \\ \bottomrule
    \end{tabular}%
    }
    \caption{Experimental results in long-tailed settings on Re-DocRED's test set.}
    \label{tab:long-tailed-result}
    \vspace{-5mm}
\end{table}

\subsection{Ablation Study}


To address \textbf{RQ3}, we conduct an ablation study on Re-DocRED (Table~\ref{tab:ablation-study}), reporting results from the best-performing models.

The Self-supervised step is critical, as removing it ("\textit{w/o ST}") leads to clear drops (dev: $-0.41/-0.29$, test: $-0.22/-0.14$ in $F_1$/Ign $F_1$).

The Disagreement-based Cross-supervision further improves performance by mitigating confirmation bias; removing it ("\textit{w/o Cross-sup}") results in additional declines (dev: $-0.14/-0.10$), highlighting the importance of both components.

\begin{table}
    \centering
    \resizebox{0.95\linewidth}{!}{
    \begin{tabular}{l c c c c}
    \toprule
    \multirow{2}{*}{\textbf{Model}} & \multicolumn{2}{c}{\textbf{Dev}} & \multicolumn{2}{c}{\textbf{Test}} \\
    \cmidrule(r){2-3} \cmidrule(r){4-5}
     & $F_1$ & Ign $F_1$ & $F_1$ & Ign $F_1$ \\
    \midrule
    \textbf{\model}               & \textbf{79.93} & \textbf{78.74} & \textbf{79.50} & \textbf{78.35} \\
    \textit{- w/o Cross-sup}      & 79.79 & 78.64 & 79.48 & 78.31 \\
    \textit{- w/o ST}             & 79.52 & 78.45 & 79.28 & 78.21 \\ \midrule
    UGDRE                         & 78.28 & 77.32 & 78.14 & 77.24 \\
    KD-DocRE                      & 76.47 & 75.30 & 76.14 & 74.97 \\
    \bottomrule
    \end{tabular}}
    \caption{Ablation study on the Re-DocRED's development set.}
    \label{tab:ablation-study}
    \vspace{-3mm}
\end{table}

\subsection{Efficiency Analysis}
To answer \textbf{RQ4}, we compare training time and data usage of \model\ against the previous state-of-the-art baselines and the performance of \model\ with different sizes of human-annotated set under the Re-DocRED dataset.

\subsubsection{Training time and Unlabelled Document Quantity}
We reported the training time and data usage of \model\ in Table~\ref{tab:train-time}.
\begin{table}[!ht]
    \centering
    \resizebox{\columnwidth}{!}{%
    \begin{tabular}{lcc} 
        \toprule 
        \textbf{Model}      & \textbf{Training Time}    & \textbf{\# of Unlabelled Documents}   \\ \midrule 
        UGDRE               & 55 hours                  & 101,873                               \\
        \textbf{\model\ (ours)}     & 26 hours                  & 30,480                                \\\bottomrule 
    \end{tabular}%
    }
    \caption{Comparison of Training Time and Unlabelled Data Usage Between \model\ and UGDRE.}
    \label{tab:train-time}
    \vspace{-3mm}
\end{table}

Our model completes training in just 26 hours, less than half the time required by UGDRE. Additionally, it achieves superior performance using only 30,480 unlabelled documents, in contrast to the 101,873 required by UGDRE. This demonstrates that our approach is significantly more resource-efficient, both in terms of computation and data consumption. Thus, \model\ is effective and resource-efficient, scaling with unlabelled data for more generalisable gains in real-world DocRE.


\subsubsection{Labelled Document Quantity}
To further demonstrate the robustness of our method, we evaluate \model\ with different sizes of human-annotated datasets during the warm-up stage, illustrated in Table \ref{tab:labelled-docs-quant}.
\begin{table}[!h]
    \centering
    \resizebox{0.95\columnwidth}{!}{%
    \begin{tabular}{lccccc}
        \toprule
        \textbf{\# of Labelled Docs} & \textbf{100} & \textbf{200} & \textbf{300} & \textbf{400} & \textbf{600} \\ \midrule
        \textbf{Dev F1}                  & 69.5         & 71.64        & 74.3         & 74.97        & 75.71        \\ \bottomrule
    \end{tabular}
    }
\caption{Performance vs. Labelled Document quantity}
\label{tab:labelled-docs-quant}
\vspace{-3mm}
\end{table}

The results show that \model\ is not highly sensitive to the size of the annotated warm-up set: with only 400 documents ($\sim$13.1\%), it achieves performance comparable to VaeDiff-DocRE \cite{tran2025vaediff}, and with 600 documents, it surpasses KD-DocRE \cite{tan2022document}. Even with 300 documents ($\sim$9.8\%), the model reaches 73.58 Dev F$_1$ during warm-up and improves to 74.28 Dev F$_1$ afterwards, demonstrating that strong performance can be achieved with minimal annotations, highlighting the practicality of our framework in low-resource DocRE scenarios.



\subsection{Result with Different Models}
\label{subsec:result-diff-models}
To answer the \textbf{RQ5}, we further conducted experiments with 2 extra models, ATLOP \cite{zhou2021document} and DocuNet \cite{zhang2021document}, alongside KD-DocRE and UGDRE and report the results in Table \ref{tab:per-diff-models}:
\begin{table}[!h]
\resizebox{\columnwidth}{!}{%
\begin{tabular}{lccc}
\toprule
\textbf{Models}                             & \textbf{Component} & \textbf{Dev F1} & \textbf{Test F1} \\ \midrule
ATLOP                                       &                    & 74.34           & 74.23            \\ 
DocuNet                                     &                    & 76.22           & 75.35            \\ 
\multirow{2}{*}{Syn2DRE (ATLOP + DocuNet)}  & \textit{ATLOP}     & 75.41           & 75.12            \\ 
                                            & \textit{DocuNet}   & 78.20           & 77.62            \\ 
Syn2DRE (ATLOP + KD-DocRE)                  & \textit{ATLOP}     & 76.18           & 75.99            \\ 
Syn2DRE (ATLOP + UGDRE)                     & \textit{ATLOP}     & 76.34           & 76.04            \\
Syn2DRE (DocuNet + KD-DocRE)                & \textit{DocuNet}   & 78.54           & 77.84            \\ 
Syn2DRE (DocuNet + UGDRE)                   & \textit{DocuNet}   & 78.13           & 77.73            \\\bottomrule
\end{tabular}%
}
\caption{Performance with different models. Results of ATLOP and DocuNet are reported from \citet{sun2023uncertainty}.}
\label{tab:per-diff-models}
\vspace{-5mm}
\end{table}

As shown in Table \ref{tab:per-diff-models}, both ATLOP and DocuNet achieve notable gains within our framework, with stronger architectures yielding better performance. The results also show that co-training with a different architecture is beneficial: ATLOP performs better when paired with the graph-based UGDRE than with the transformer-based KD-DocRE, while DocuNet shows the opposite trend.

\section{Related Works}
\subsection{Document-level Relation Extraction.}
\label{subsec:docre-models-related-works}
Document-level Relation Extraction (DocRE) identifies relations between entity pairs across sentences or entire documents. Existing methods mainly fall into two categories: \textit{transformer-based} and \textit{graph-based}. Transformer-based approaches \citep{zhou2021document, tan2022document, zhang2021document} enhance entity representations using attention mechanisms in pre-trained language models, including localised context embeddings \citep{zhang2021document} and multi-hop reasoning via Axial Attention \citep{tan2022document}. In contrast, graph-based methods \citep{christopoulou2019connecting, zeng2020double, xu2021document} model document structure explicitly by constructing heterogeneous graphs (e.g., mention, entity, and sentence nodes) and applying graph-based reasoning to improve relation prediction.

\subsection{Synthetic Data for DocRE.}
Aforementioned DocRE systems in Section \ref{subsec:docre-models-related-works} typically require large amounts of high-quality labelled data, which is costly and prone to annotation errors. To mitigate this, recent work explores synthetic data to improve scalability. One direction focuses on data generation \citep{tan2022document, sun2023uncertainty, feng2024dure, tran2025vaediff}, such as generating pseudo-texts for long-tail relations \citep{feng2024dure} or synthesising entity pair representations \citep{tran2025vaediff}. Another direction leverages pseudo-labelling, where teacher models annotate unlabelled data for student training \citep{tan2022document}, with extensions incorporating uncertainty estimation via Monte Carlo dropout and multi-stage training \citep{sun2023uncertainty}.

However, these methods still have notable limitations. \citet{tran2025vaediff}'s method continues to rely on substantial amounts of high-quality labelled data for strong performance. Meanwhile, methods proposed by \citet{sun2023uncertainty, feng2024dure} are computationally expensive due to the large number of documents and entity pairs. For instance, \citet{feng2024dure} generate pseudo-documents for each entity pair, while \citet{sun2023uncertainty} require multiple rounds of training and denoising. Consequently, \model\ is proposed to reduce computational cost while producing more reliable pseudo-labels than prior methods.
\section{Conclusion}
In conclusion, we introduced \model, a semi-supervised framework that effectively exploits unlabelled and weakly labelled data to advance DocRE. To reduce training overhead, we employed a lightweight Self-supervised strategy that generates reliable pseudo-labels on the fly. To combat confirmation bias, we proposed a Cross-supervision mechanism that jointly trains two teacher-student networks in a collaborative manner. These components are integrated into a unified, end-to-end training procedure. Experimental results and ablation studies confirm that \model\ achieves state-of-the-art performance with improved efficiency.
\newpage
\section*{Limitations}
\label{sec:limitation}
First, \model\ still requires a non-trivial amount of labelled data, limiting its applicability in low-resource domains such as biomedicine; future work could improve performance under limited supervision or explore leveraging LLMs for pseudo-label generation. Second, its dual-network design introduces substantial GPU memory overhead, nearly doubling usage compared to UGDRE, motivating more memory-efficient alternatives. Third, the method is restricted to a binary dual-system setup, without exploring extensions to more than two models, which remains for future work. Finally, pseudo-label thresholds are fixed in this paper, and learning adaptive thresholds is left for future work.
\bibliography{reference}

\appendix
\label{sec:appendix}
\section{Baseline Structures}
\label{appendix:baseline-structures}

\subsection{UGDRE}
\label{sec:ugdre}
\paragraph{Architecture.}
We briefly describe the architecture of the UGDRE model proposed by \citet{sun2023uncertainty}.

UGDRE begins by constructing a document-level graph to model non-local dependencies among entities. In this graph, each node corresponds to a token in the document. Edges are established based on the following heuristic rules: (1) tokens within the same mention are connected to preserve mention semantics, (2) tokens from different mentions of the same entity are connected to model cross-mention interactions, and (3) tokens of entities co-occurring in the same sentence are connected to capture sentence-level interactions.

Graph inference is performed using Graph Convolutional Networks (GCNs), defined as:
\begin{equation} 
    g_i = \rho\left(\sum_{j=1}^n A_{ij} W z_j + b\right) 
\end{equation} 
where $W \in \mathbb{R}^{d \times d}$ and $b \in \mathbb{R}^d$ are learnable parameters, $z_j$ is the contextual representation of the $j$-th token, $A_{ij}$ is the document graph’s adjacency matrix, and $\rho$ is a non-linear activation function.

Each token’s final representation $h_i$ is obtained by concatenating its contextual and graph-enhanced embeddings: 
\begin{equation} 
    h_i = [z_i, g_i] 
\end{equation}

Entity representations ${e_i}_{i=1}^q$ are computed via \texttt{logsumexp} pooling over token representations. To predict the relation between a pair of entities $(e_i, e_j)$, UGDRE applies a group bilinear operator to generate the probability distribution over all relation types.

\paragraph{Pre-training Phase Hyperparameters.}
Table \ref{tab:hyper-params-pretrain-ugdre} represents the hyper-parameters used when training UGDRE during Pre-training Phase (Section \ref{sec:pre-training-stage}).
\begin{table}[!th]
  \centering
  \begin{tabular}{lc}
    \toprule
    \textbf{Hyper-parameters}           & \textbf{Value}\\
    \midrule
    Train batch size                    & 8             \\
    Test batch size                     & 8             \\
    Dropout rate                        & 0.25          \\
    Warmup ratiors                      & 6\%           \\
    Train epochs                        & 10            \\ 
    Encoder learning rate               & $5e-5$        \\
    Classifier learning rate            & $1e-4$        \\
    Maximum gradient norm               & 1.0           \\  \bottomrule
  \end{tabular}
  \caption{Hyper-parameter settings for pre-training UGDRE \citep{sun2023uncertainty}.}
  \label{tab:hyper-params-pretrain-ugdre}
\end{table}

\subsection{KD-DocRE}
\label{sec:kddocre}
\paragraph{Architecture.}
We provide an overview of the KD-DocRE model proposed by \citet{tan2022document}.

To highlight entity mentions, KD-DocRE inserts a special token ``*'' before and after each mention $m_j^i$ of an entity $e_i$, and encodes the full document using a pre-trained language model. Entity representations are then computed via \texttt{logsumexp} pooling over the mention representations, where each mention is represented by the embedding of its starting ``*'' token. The representation of an entity pair $(e_h, e_t)$ is computed as follows:
\begin{equation}
    \boldsymbol{p}_{h,t} = \text{tanh}\left(\boldsymbol{W}_p[\boldsymbol{e}_h;\boldsymbol{e}_t;\boldsymbol{c}_{h,t}] + \boldsymbol{b}_p\right)
\end{equation}
where $\boldsymbol{W}_p$ and $\boldsymbol{b}_p$ are trainable parameters, $[\,;\,]$ denotes concatenation, and $\boldsymbol{c}_{h,t}$ is the localised context embedding for the entity pair \citep{zhou2021document}.

To capture structured interactions among entities, KD-DocRE introduces Axial Attention \citep{ho2019axial}. Specifically, an entity-level table is constructed where each row and column corresponds to an entity in the document. Self-attention \citep{vaswani2017attention} is then applied independently along both the row (height) and column (width) axes to model inter-entity dependencies.

For training, \citet{tan2022document} propose an enhanced version of Adaptive Thresholding Loss \citep{zhou2021document} by incorporating Focal Loss \citep{lin2017focal}:
\begin{equation}
    \mathcal{L} = \sum_{r_i \in \mathcal{P}_T}\left( 1 - P(r_i)\right)^\gamma \log\left(P(r_i)\right) + \log\left(P(r_{\text{TH}})\right)
\end{equation}
where $\mathcal{P}_T$ denotes the set of positive relations for the entity pair $(e_h, e_t)$, $P(r_i)$ is the predicted probability for relation $r_i$, and $P(r_{\text{TH}})$ is the predicted probability for the threshold relation \textit{TH}.
 
\paragraph{Pre-training Phase Hyperparameters.}
Table \ref{tab:hyper-params-pretrain-kddocre} represents the hyper-parameters used when training KD-DocRE during Pre-training Phase (Section \ref{sec:pre-training-stage}).
\begin{table}[!h]
  \centering
  \begin{tabular}{lc}
    \toprule
    \textbf{Hyper-parameters}           & \textbf{Value}\\
    \midrule
    Train batch size                    & 8             \\
    Test batch size                     & 8             \\
    $\gamma$ in AFL loss function       & 0.5           \\
    Warmup ratiors                      & 6\%           \\
    Train epochs                        & 5             \\ 
    Encoder learning rate               & $3e-5$        \\
    Classifier learning rate            & $1e-4$        \\
    Maximum gradient norm               & 1.0           \\  \bottomrule
  \end{tabular}
  \caption{Hyper-parameter settings for pre-training KD-DocRE \citep{tan2022document}.}
  \label{tab:hyper-params-pretrain-kddocre}
\end{table}

\section{Experimental Details}
\label{appendix:experimental-details}
\subsection{Dataset Statistics}
\label{appendix:dataset-statistics}
Table \ref{tab:datasets-statistics} provides the statistics of datasets used in this paper. All datasets are publicly available.
\begin{table}[!ht]
  \centering
  \resizebox{\columnwidth}{!}{
    \begin{tabular}{l c c}
        \toprule
        \textbf{Datasets}   & \textbf{\# Documents}     \\
        \midrule
        DocRED-Train                & 3,053                     \\
        DocRED-Dev                  & 1,000                     \\
        DocRED-Test                 & 1,000                     \\
        DocRED Distantly Supervised        & 101,873            \\ \midrule
        Re-DocRED-Train             & 3,053                     \\
        Re-DocRED-Dev               & 500                       \\ 
        Re-DocRED-Test              & 500                       \\ \midrule
        HacRED-Train                & 6,231                     \\
        HacRED-Dev                  & 1,500                     \\
        HacRED-Test                 & 1,500                     \\ \midrule
        CDR-Train                   & 500                       \\
        CDR-Dev                     & 500                       \\
        CDR-Test                    & 500                       \\ \midrule
        GDA-Train                   & 23,353                    \\
        GDA-Dev                     & 5,839                     \\
        GDA-Test                    & 1,000                     \\ \bottomrule
      \end{tabular}
  }
  \caption{Statistics of the datasets in experiments.}
  \label{tab:datasets-statistics}
\end{table}

\section{Derivation of Eq. (\ref{ema-backprop})}
\label{sec:eq-5-derivation}
In this section, we detail how the EMA update can be reorganised in the form of back-propagation. Equation (\ref{eq:student-params-updates}) defines the update rule for the student parameters $\theta_s$ using standard back-propagation at the $j$-th iteration, while Equation (\ref{eq:ema-teacher-update-appendix}) presents the EMA update for the teacher parameters $\theta_t$.

\begin{equation}
    \label{eq:student-params-updates}
    \theta_s^j = \theta_s^{j-1} - \gamma \cdot \pdv{\mathcal{L}}{\theta_s^{j-1}}
\end{equation}

\begin{equation}
    \label{eq:ema-teacher-update-appendix}
    \theta_t^j = \alpha \cdot \theta_t^{j-1} + (1-\alpha) \cdot \theta_s^j
\end{equation}

By substituting Equation (\ref{eq:student-params-updates}) into Equation (\ref{eq:ema-teacher-update-appendix}), we can express the teacher parameter $\theta_t$ at iteration $j$ as:
{
\footnotesize
\begin{align}
\theta^j_t &= \alpha \cdot \theta^{j-1}_t + (1-\alpha) \cdot \theta^{j}_s \nonumber\\
&= \hspace{1mm} \alpha^{j} \cdot \theta^{0}_t + \alpha^{j-1}(1-\alpha)\left(\theta^{0}_s - \gamma \cdot  \frac{\partial \mathcal{L}}{\partial \theta^{0}_s}\right) + \dots \\ 
& \hspace{5mm} + (1-\alpha)\left(\theta^{j-1}_s - \gamma \cdot \frac{\partial \mathcal{L}}{\partial \theta^{j-1}_s}\right) \nonumber \\
&= \hspace{1mm} \alpha^{j} \cdot \theta^{0}_t + (1-\alpha) \sum_{i=0}^{j-1}\alpha^{i} \cdot \theta^{0}_s  \\
& \hspace{5mm} - \gamma(1-\alpha) \sum_{i=0}^{j-1}\sum_{k=0}^{j-1-i} \alpha^{k} \cdot \frac{\partial \mathcal{L}}{\partial \theta_s^i} \nonumber\\
&= \alpha^j \cdot \theta^0_t + \theta^0_s - \alpha^j \cdot \theta^0_s - \gamma \sum_{i=0}^{j-1}\left(1 - \alpha^{j-i}\right) \cdot \frac{\partial \mathcal{L}}{\partial \theta_s^i} \nonumber\\
&= \theta^0_s - \gamma \sum_{i=0}^{j-1}\left(1 - \alpha^{j-i}\right) \cdot \frac{\partial \mathcal{L}}{\partial \theta_s^i} \nonumber\\
&= \bar{\theta} - \gamma \sum_{i=0}^{j-1}\left(1 - \alpha^{j-i}\right) \cdot \frac{\partial \mathcal{L}}{\partial \theta_s^i} \nonumber\\
&\text{where} \quad \theta^0_s = \theta^0_t = \bar{\theta}
\end{align}
}

Thus, the teacher parameter at the $(j{-}1)$-th iteration becomes:
\begin{equation}
\theta^{j-1}_t = \bar{\theta} - \gamma \sum_{i=0}^{j-2}\left(1 - \alpha^{j-1-i}\right) \cdot \frac{\partial \mathcal{L}}{\partial \theta_s^i}
\end{equation}

To express the EMA update in the form of back-propagation, consider the gradient step:

$$
\theta_t^j = \theta_t^{j-1} - \nabla
$$

Then,
\begin{align}
\nabla &= \theta_t^{j-1} - \theta_t^j \nonumber \\
&= \hspace{1mm} \gamma \sum_{i=0}^{j-1}\left(1 - \alpha^{j-i}\right) \cdot \frac{\partial \mathcal{L}}{\partial \theta_s^i} \\ 
& \hspace{5mm} - \gamma \sum_{i=0}^{j-2}\left(1 - \alpha^{j-1-i}\right) \cdot \frac{\partial \mathcal{L}}{\partial \theta_s^i} \nonumber\\
&= \gamma (1 - \alpha) \sum_{i=0}^{j-1} \alpha^{j-1-i} \cdot \frac{\partial \mathcal{L}}{\partial \theta_s^i}
\end{align}

Substituting back, we obtain the back-propagation form of the EMA update:
\begin{equation}
\theta^j_t = \theta^{j-1}_t - \gamma (1 - \alpha) \sum_{i=0}^{j-1} \alpha^{j-1-i} \cdot \frac{\partial \mathcal{L}}{\partial \theta_s^i}
\end{equation}

Here, $\gamma$ is the learning rate and $(1 - \alpha)$ is typically a small value, as $\alpha$ is generally close to 1 (e.g., 0.995). This formulation reveals that the EMA update corresponds to applying a scaled, weighted sum of historical gradients from the student model. The conservative and smoothing nature of this update makes EMA particularly effective for producing stable and reliable teacher models.



\subsubsection{Hardware and software configurations}
\label{sec:hardware-software-config}
The models were implemented using PyTorch \citep{paszke2019pytorch} and Huggingface Transformers \citep{wolf2020transformers}. For EMA\footnote{\href{https://github.com/lucidrains/ema-pytorch}{https://github.com/lucidrains/ema-pytorch}} and Axial Attention\footnote{\href{https://github.com/lucidrains/axial-attention}{https://github.com/lucidrains/axial-attention}}, we adopted publicly available repositories to construct teacher networks and implement KD-DocRE. Full software configuration details are provided in Table~\ref{tab:package-config}.
\begin{table}[!ht]
  \centering
  \resizebox{0.6\linewidth}{!}{
    \begin{tabular}{l c}
        \toprule
        \textbf{Package}            & \textbf{Version} \\
        \midrule
        axial\_attention            & 0.6.1       \\
        ema\_pytorch                & 0.5.2       \\
        numpy                       & 1.26.2       \\
        opt\_einsum                 & 3.3.0       \\
        PyTorch                     & 2.1.2         \\ 
        tqdm                        & 4.66.1         \\
        Transformers                & 4.32.1     \\  \bottomrule
      \end{tabular}
  }
  \caption{Package configurations of our experiments.}
  \label{tab:package-config}
\end{table}

\end{document}